\documentclass[10pt,twocolumn,letterpaper]{article}

\usepackage{iccv}
\usepackage{times}
\usepackage{epsfig}
\usepackage{graphicx}
\usepackage{amsmath}
\usepackage{amssymb}

\usepackage{booktabs}
\usepackage{multirow}
\usepackage{colortbl}
\usepackage{xcolor}
\usepackage{subfig}
\usepackage{makecell}
\usepackage{algorithm}
\usepackage{algorithmicx}
\usepackage[noend]{algpseudocode}

\newenvironment{sequation}{\small\begin{equation}}{\end{equation}}

\usepackage{pifont}

\usepackage{amsthm}
\newtheorem{thm}{Theorem}

\usepackage[breaklinks=true,bookmarks=false]{hyperref}

\iccvfinalcopy 

\def\iccvPaperID{6986} 

\ificcvfinal\fi

\begin{document}

\title{CBA: Improving Online Continual Learning via Continual Bias Adaptor}

\author{
Quanziang Wang\textsuperscript{1}
\and Renzhen Wang\textsuperscript{1*}
\and Yichen Wu\textsuperscript{2}
\and Xixi Jia\textsuperscript{3}
\and Deyu Meng\textsuperscript{1,4}\thanks{Corresponding author} 
\and \textsuperscript{1} Xi'an~Jiaotong~University
\and \textsuperscript{2} City~University~of~Hong~Kong
\and \textsuperscript{3} Xidian~University
\and \textsuperscript{4} Macau~University~of~Science~and~Technology \\
{\tt\small quanziangwang@gmail.com, \{rzwang, dymeng\}@xjtu.edu.cn}
}

\maketitle
\ificcvfinal\thispagestyle{empty}\fi

\begin{abstract}
   Online continual learning (CL) aims to learn new knowledge and consolidate previously learned knowledge from non-stationary data streams. Due to the time-varying training setting, the model learned from a changing distribution easily forgets the previously learned knowledge and biases toward the newly received task. To address this problem, we propose a Continual Bias Adaptor (CBA) module to augment the classifier network to adapt to catastrophic distribution change during training, such that the classifier network is able to learn a stable consolidation of previously learned tasks. In the testing stage, CBA can be removed which introduces no additional computation cost and memory overhead. We theoretically reveal the reason why the proposed method can effectively alleviate catastrophic distribution shifts, and empirically demonstrate its effectiveness through extensive experiments based on four rehearsal-based baselines and three public continual learning benchmarks\footnote{Code is available at \href{https://github.com/wqza/CBA-online-CL}{\textcolor{magenta!80}{https://github.com/wqza/CBA-online-CL}}}.
\end{abstract}

\section{Introduction}
\label{sec:introduction}
Continual learning (CL)~\cite{cl-settings, cl-survey} focuses on designing a model that can learn continuously from streaming data and accumulate new knowledge while consolidating previously learned knowledge. In the context of CL, the data distribution of streaming tasks is in general non-stationary and changes over time, which violates the independent and identically distributed (i.i.d) assumption that is commonly adopted in machine learning. Therefore, continual learning suffers from the notorious catastrophic forgetting problem~\cite{forgetting}, where the model severely forgets the previously learned knowledge after being trained on a new task. 

Traditional offline CL stores all training batches of the current task and the model is trained on these samples for multiple epochs with repeat shuffle. However, the availability of previously learned batches might be restricted due to privacy concerns~\cite{cl-survey2} or memory limitations. In this paper, we mainly focus on a more challenging and realistic setting, online CL~\cite{online-survey}, where samples from each task can be trained only single-pass (\textit{i.e.}, one epoch) and the past batches are not accessible. 

In online CL, the distribution of the training data changes over time and it differs not only from the joint distribution of all tasks (as in offline CL) but also from the distribution of the task they belong to. Therefore, online CL commonly causes an even more severe distribution shift, which further intensifies catastrophic forgetting. To alleviate this problem, rehearsal-based algorithms~\cite{icarl, cl-survey} employed a small memory buffer to store examples of previous tasks so as to approximate the joint distribution of all seen training data, then collectively trained the model on the memory buffer with the current task. Along this line, DER++~\cite{DER} utilized an additional knowledge distillation to further reply logits of old task samples, and RAR~\cite{RAR} used random augmentation to address the overfitting problem of the small memory buffer. In another vein, a wide range of works attribute catastrophic forgetting to task-recency bias \cite{SCR}, \textit{i.e.}, the classifiers tend to classify samples into currently received classes. They in turn proposed to improve the original linear classifier \cite{SSIL, LUCIR, bic} or directly replace the classifier with the nearest classifier \cite{SCR, icarl} to mitigate the adverse effects of class imbalance between currently received classes and replayed classes. Despite the promising performance, almost all of these methods implicitly view task-recency bias as a label distribution shift and tackle it from the perspective of class imbalance problem, which makes these methods sub-optimal in practice \cite{online-er-ace}.

\begin{figure}
\centering
\includegraphics[width=0.9\columnwidth]{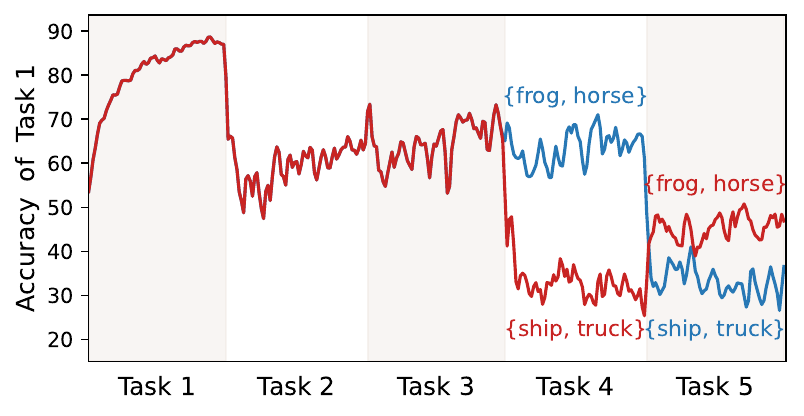}
\caption{Comparing the accuracy change of the categories learned in 1st task. For the 4th and 5th tasks, the \textbf{red} corresponds to learning  from $\{ship, truck\}$ to $\{frog, horse\}$, and the \textbf{blue} corresponds to learning from $\{frog, horse\}$ to $\{ship, truck\}$.
}
\label{fig:discrepancy}
\vspace{-2mm}
\end{figure}

In fact, the target of online CL is to accomplish a stable consolidation of knowledge for all learned tasks, by training a classifier to fit the posterior probability $\mathbb P(Y|X)$, where $X$ and $Y$ represent stochastic variables of the input data and the corresponding label, respectively.
According to Bayes's rule $\mathbb P(Y|X) \propto \mathbb P(X|Y)\mathbb P(Y)$, the posterior probability depends on both the prior probability $\mathbb P(Y)$ and the likelihood $\mathbb P(X|Y)$.
It can be seen that any shift from $\mathbb P(Y)$ or $\mathbb P(X|Y)$ will lead to distribution change in $\mathbb P(Y|X)$.
On the one hand, as aforementioned, the label distribution $\mathbb P(Y)$ shifts leading to severe forgetting when a new task comes in.
On the other hand, $\mathbb P(X|Y)$ can also suffer from catastrophic shifts \textcolor{black}{(dubbed feature distribution shift for simplicity)} due to the time-varying data streams.
To illustrate this, we conduct a toy experiment with Experience Replay (ER)~\cite{ER1} on CIFAR-10~\cite{cifar}, where we manipulate the incoming categories of the 4th task (represented by the red and blue lines, respectively), and keep tracking the accuracy of the categories of the 1st task as shown in Fig.\ref{fig:discrepancy}.
Notably, $\mathbb P(X|Y)$ of the two lines are totally different, while their label distribution $\mathbb P(Y)$ maintains the same tendency. This disparity leads to a catastrophic change in the performance of categories learning in the 1st task.
This validates the existence of feature distribution shifts and poses a challenging problem for online CL: how to accomplish a stable consolidation of past knowledge under these distribution shifts.

To tackle this challenge, based on rehearsal-based methods, this paper proposes a bi-level learning framework to adapt the posterior distribution shift during each training step directly. Specifically, we introduce a nonlinear transformation module, Continual Bias Adaptor (CBA), to dynamically capture the catastrophic distribution change for posterior probability $\mathbb P(Y|X)$. Simultaneously, the original classifier is adjusted to fit an implicit posterior probability which tends to be a stable consolidation of previously learned knowledge across different tasks. 
Intrinsically, the proposed method implicitly aligns the distribution between training and memory buffer data at each iteration.
During the testing stage, the CBA module can be simply removed and the original classifier can achieve a good performance on all seen tasks. To summarize, our main contributions are three-fold: 
\begin{itemize}
    \item We propose a bi-level learning framework to model the posterior distribution shift in an online manner. We theoretically investigate the proposed method from gradient alignment and reveal the reason why it can effectively alleviate catastrophic distribution shifts.
    \item We propose a CBA module that can plug in most of the rehearsal-based methods during training, and be removed in the test stage so that it involves no computation cost and memory overhead in inference.
    \item We evaluate the performance based on four rehearsal-based baselines with extensive experiments over various benchmarks. We show that the proposed method can effectively consolidate previously learned knowledge. This is also demonstrated by task-blurry online CL and offline CL settings.
\end{itemize}

\section{Related Work}
\label{sec:related-work}

\noindent\textbf{Continual learning settings.} Based on different task construction manners, continual learning mainly falls into three categories~\cite{cl-settings, cl-survey}: Task-incremental learning (Task-IL), Domain-incremental learning (Domain-IL), and Class-incremental learning (Class-IL). Specifically, Task-IL assumes the task identity is provided, which means the model can use the task index in both the training and testing stage. Domain-IL focuses on the concept drift where the domain of each task is changing but with the same label space~\cite{domain-il3, domain-il2, domain-il1}. This paper concentrates on the more challenging Class-IL, where the task indices are absent during testing~\cite{DER, class-il1, icarl, class-il2}. 

From the training perspective, CL can be divided into offline and online CL. Offline CL could preserve all samples of the current task, permitting repeated training on them~\cite{CLSER, DER, e2e, offline1, icarl}. As for online CL, samples of each task cannot be stored and each sample can only be seen once, except for those examples saved in the memory buffer~\cite{ACC_AUC, online-survey}. In this paper, we mainly focus on online CL, which is more demanding and realistic than offline CL.

\medskip
\noindent\textbf{Rehearsal-based methods in online CL:}
The objective of online CL is to learn models with a stronger ability to quickly fit new tasks while retaining knowledge of old tasks~\cite{online-er-ace, online-imbalance, ACC_AUC, RER, OCS}.
Experience replay (ER), the most commonly used baseline, trains the new incoming samples together with old samples stored in the memory buffer. Its variants attempt to combine replay strategies with other techniques, such as knowledge distillation and random augmentation. For example, DER++~\cite{DER} utilized the knowledge distillation technique to further replay logits of memory buffer data. RAR~\cite{RAR} adopted random augmentation to alleviate the overfitting of the tiny memory buffer, and CLSER~\cite{CLSER} constructed a plastic model and a stable model for recent and structural knowledge distillation. Different from these methods, some studies emphasize maximizing the utility and benefits of the memory buffer samples. Instead of randomly sampling, GSS~\cite{gradient-selection-blurry} selected the samples stored in the memory buffer according to the cosine similarity of gradients. MIR~\cite{MIR} chose maximally interfering samples whose prediction will be most negatively impacted by the foreseen parameters update. OCS~\cite{OCS} picked the most representative data of the current task while minimizing interference to previous tasks. Unlike these methods, our method focuses on alleviating distribution shifts and can plug in most current rehearsal-based approaches.

\medskip
\noindent\textbf{Task-recency bias in online CL:}
Task-recency bias \cite{LUCIR, cl-survey} in online CL means that classifiers tend to mistakenly classify previously learned classes as newly encountered ones. A popular perspective is that the linear classifier is particularly susceptible to task-recency bias. To address this, iCaRL~\cite{icarl} proposed to replace the linear classifier with nearest class mean (NCM) classifiers. Similarly, SCR~\cite{SCR} and Co$^2$L~\cite{Co2L} employed the NCM classifier where the feature extractor is learned from contrastive learning. A wide range of works tackles the task-recency bias as a class imbalance problem~\cite{SSIL, e2e, LUCIR, bic, WA}. For example, LUCIR~\cite{LUCIR} added weight normalization on the linear classifier. BiC~\cite{bic} proposed a bias correction layer turned on a held-out validation set. SS-IL~\cite{SSIL} separated the softmax to mitigate the imbalanced penalization of the old class outputs. On the flip side, ER-ACE~\cite{online-er-ace} pointed out that task-recency bias can also arise from feature interference and designed an asymmetric loss to address this problem. Different from these methods, our proposed method relaxes the assumption on label/feature distribution shift by directly modeling the posterior distribution shift.
Besides, some methods addressing class imbalance may potentially be explored for tackling task-recency bias in online CL~\cite{autobalance, logit_adjustment, meta_modulator, L2AC}. However, many of these approaches face challenges in generalization to CL due to unstable data distribution.

\section{Method}
\label{sec:method}

\begin{figure*}[t]
\begin{center}
   \includegraphics[width=0.8\linewidth]{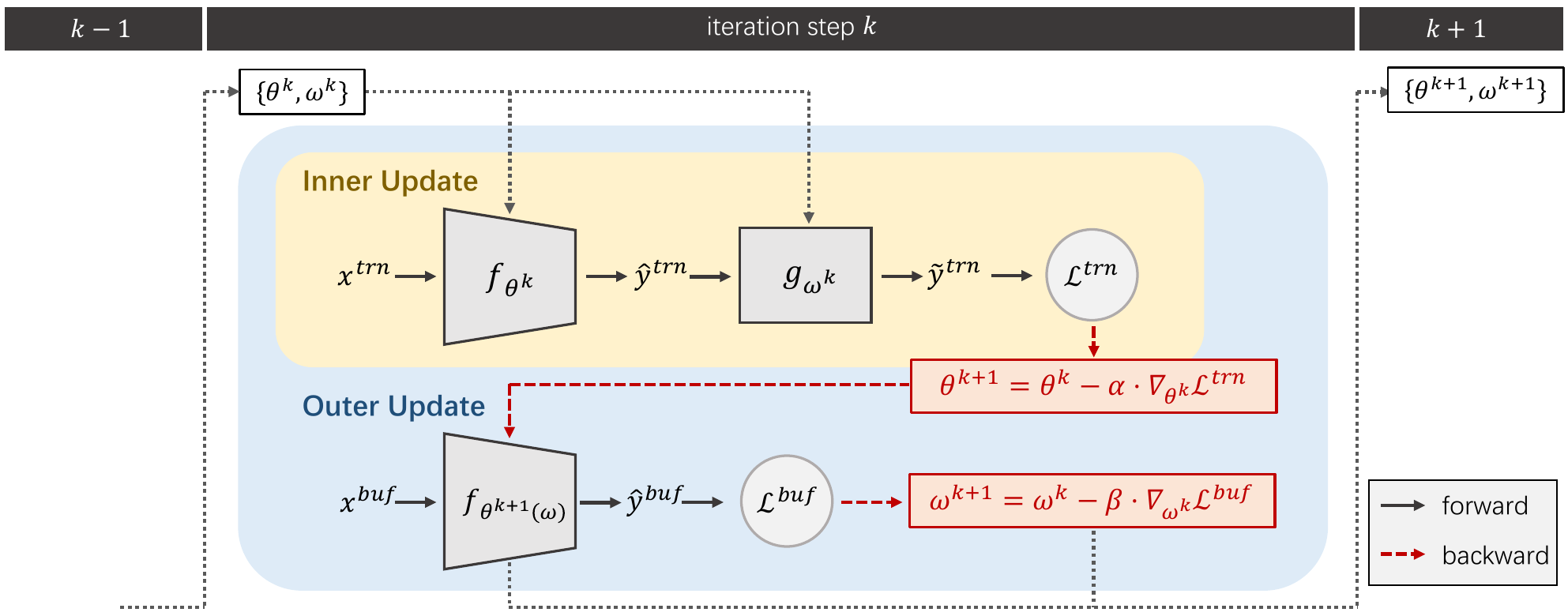}
\end{center}
\caption{Method overview. At the iteration step $k$, for the inner loop, the forward process computes the training loss in Eq.~(\ref{eq:inner-loss}) and the backward process updates the classification model parameter $\theta(\omega)$ by Eq.~(\ref{eq:inner-opt}). And for the outer loop, the forward process computes the loss in Eq.~(\ref{eq:outer-loss}) and the backward process updates the CBA parameter $\omega$ by Eq.~(\ref{eq:outer-opt})}
\label{fig:main}
\end{figure*}

\subsection{Preliminaries}
\label{subsec:preliminaries}
Suppose we have $N$ sequential tasks $\{\tau_1, \tau_2, \cdots, \tau_N\}$, each task $\tau_i=\{(x^i_j,y^i_j)\}_{j=0}^{N_i}$, where $N_i$ is the total number of training samples in $\tau_i$. In online CL, the classification model $f_\theta$ can only train single-pass on samples of each task $\tau_i$ (\textit{i.e.}, train with one epoch), and the previously seen batches are not accessible. 
Let $\tau_t$ represent the current learning task, $\mathcal{M}$ denotes the tiny memory buffer which stores the samples of previous tasks $\{\tau_{1}, \cdots, \tau_{t-1}\}$. The most representative rehearsal-based method Experience Replay (ER)~\cite{ER1, ER2} jointly trains current task $\tau_t$ with examples sampled from the buffer $\mathcal{M}$, and its training objective function $\mathcal{L}$ is:
\begin{equation}
    \mathcal{L}^{trn} \left(\mathcal{B}^{trn}; f_\theta \right) = 
    \frac{1}{|\mathcal{B}^{trn}|}
    \sum_{x,y \in \mathcal{B}^{trn}} L(f_\theta(x), y) ,
\end{equation}
where the training batch $\mathcal{B}^{trn}=\mathcal{B}^t \cup \mathcal{B}^{buf}$ consists of a batch of incoming new samples 
 $\mathcal{B}^t \subset \tau_t$ and a batch sampling from the memory buffer $\mathcal{B}^{buf} \subset \mathcal{M}$, and $L$ denotes the cross-entropy loss.

Note that the memory buffer $\mathcal{M}$ is updated by reservoir sampling after training each batch $\mathcal{B}_{t}$, which means $\mathcal{M}$ not only includes samples of previous tasks but also contains the samples of the current task $\tau_t$. Therefore, the memory buffer $\mathcal{M}$ is a relatively balanced set containing samples of all seen tasks.

\subsection{Framework Formulation}
Our focus in this work is to alleviate catastrophic forgetting of rehearsal-based models under a challenging online CL setting. As aforementioned, current rehearsal-based models commonly suffer from catastrophic distribution changes due to time-varying data streams. 
To address the problem, we relax the assumption on the label or feature distribution shift in comparison to prior methods and propose to directly model the catastrophic distribution change for posterior probability $\mathbb P(Y|X)$, making the original $f_\theta$ learn a stable consolidation of knowledge for all past tasks.

The main methodology is to design a Continual Bias Adaptor (CBA) $g_\omega$ that: 1) dynamically augments the classifier network $f_\theta$ to produce more diverse posterior distributions by only tuning the parameters of $g_\omega$ ($\omega$ can be viewed as hyper-parameters) to adapt to catastrophic posterior change; 2) makes the original classifier $f_\theta$ fit an implicit posterior that tends to achieve a stable consolidation of knowledge of previously learned tasks. For a training example $x^{trn}$, its posterior probability is adjusted by the augmented classifier network $\mathcal F_{\theta,\omega}=g_\omega\circ f_\theta$ in an online manner, which can be formulated as
\begin{equation}
    \tilde{y}^{trn} = \mathcal F_{\theta,\omega}(x^{trn})= g_{\omega} \circ f_{\theta}(x^{trn}),
\label{eq:ba_output}
\end{equation}
where $\circ$ is function composition operator, and $g_\omega$ is designed as a light-weight network. 
Thus, the augmented classifier network $\mathcal F_{\theta,\omega}$ is firstly updated to learn new knowledge from the training data $\mathcal B^{trn}=\mathcal B^t \cup \mathcal B^{buf}$ that minimizes the rehearsal-based empirical risk, \textit{i.e.},
\begin{equation}
    \theta^*(\omega)=\arg\min_\theta \mathcal{L}^{trn} \left(\mathcal{B}^{trn}; \mathcal F_{\theta, \omega} \right),
\label{eq:inner-loss}
\end{equation}
where $\omega$ is actually a hyper-parameter of the optimal $\theta^*$. Note that the training loss function $\mathcal{L}^{trn}$ can be adopted by different rehearsal-based loss functions like ER and its variants such as DER++, and CLSER. Here we take the ER formulation as an example for simplicity.

On the other hand, our ultimate objective is to protect the original classifier network $f_\theta$ from catastrophic distribution shift, while accomplishing a stable consolidation of knowledge across different tasks.
To this end, we further keep tracking the performance of the classifier network to prevent catastrophic forgetting, which requires that $f_{\theta^*(\omega)}$ returned by minimizing the rehearsal-based empirical risk Eq.~(\ref{eq:inner-loss}), should maximize the performance of previously seen data.
Since accessing all of this historical data is unfeasible, it can be approximated by the empirical risk over the memory buffer data, \textit{i.e.},

\begin{sequation}
\begin{aligned}
    \omega^* &\triangleq \arg\min_\omega \mathcal{L}^{buf} \left(\mathcal{B}^{buf}; f_{\theta^*(\omega)} \right), \\
             &= \arg\min_\omega \frac{1}{|\mathcal{B}^{buf}|} \sum_{x,y \in \mathcal{B}^{buf}} L\left(f_{\theta^*(\omega)}(x), y\right).
\label{eq:outer-loss}
\end{aligned}
\end{sequation}
This formulation attempts to find the optimal parameters such that the classifier network returned by optimizing Eq.~(\ref{eq:inner-loss}), should also have a good performance on the memory buffer data which acts as a stable consolidation of knowledge from the learned tasks.

In fact, Eq.~(\ref{eq:inner-loss}) and Eq.~(\ref{eq:outer-loss}) formulate a bi-level learning framework. In the inner loop Eq.~(\ref{eq:inner-loss}), the augmented classifier network $F_{\theta,\omega}$ is updated to learn new knowledge and rehearse old knowledge from $\mathcal B^{trn}$. In the outer loop Eq.~(\ref{eq:outer-loss}), the parameters of $g_\omega$ are updated from $\mathcal B^{buf}$ to consolidate the previously learned knowledge, which has been updated in the inner loop, against catastrophic posterior change.

We herein detail the proposed CBA module and bi-level optimization algorithm in the following aspects:

\medskip
\noindent\textbf{Continual Bias Adaptor:}
The proposed CBA module $g_\omega$ is designed to be a nonlinear transformation that takes the outputs (logits) of the original classifier network $f_\theta$ as its inputs. As such, the original classifier network $f_\theta$ is augmented by cascading the CBA module after its classification layer, as shown in Fig.~\ref{fig:main}. To capture catastrophic distribution change during continual learning, we parameterize $g_\omega$ as a multi-layer perceptron (MLP) network with a single hidden layer containing 256 nodes. Each hidden node is equipped with a ReLU \cite{relu} activation function, and the output employs the Softmax activation function to guarantee a posterior probability output. Albeit simple, this network is known as a universal approximator, capable of fitting almost any continuous function \cite{MLP-universal-approx} and thus can fit various posterior distribution changes. Additionally, we introduce a skip-layer connection within the CBA module, connecting the outputs of $f_\theta$ to the MLP outputs, which aids the model convergence and facilitates the gradient backward propagation, which has been verified in previous works \cite{resnet, densenet}.

Note that the proposed CBA module is only used in the training stage. In the test stage, the test sample $x^{tst}$ is predicted by $f_\theta$, that is $\hat{y}^{tst} = f_\theta(x^{tst})$. This indicates that our method does not introduce any calculation overhead in the test stage. As a result, our method can be inferred at any time in the learning process of online CL~\cite{ACC_AUC}.

\medskip
\noindent\textbf{The learning of CBA.}
The Eq.~(\ref{eq:inner-loss}) and Eq.~(\ref{eq:outer-loss}) in the bi-level optimization are nested with each other.
Concretely, the solution $\theta^*(\omega)$ of Eq.~(\ref{eq:inner-loss}) depends on the hyper-parameter $\omega$ and the optimum solution is obtained at $\omega^*$.
However, the required optimum $\omega^*$ is solved by Eq.~(\ref{eq:outer-loss}) which relies on the best $\theta^*(\omega)$.
Indeed, there is generally no closed-form solution to the bi-level optimization framework~\cite{bilevel} and we approximately update the $\theta$ and $\omega$ using a gradient-optimization-based method following~\cite{learn2reweight, meta-weight-net}.

(1) \textbf{Update $\theta$.}
Given the CBA parameter $\omega^k$ at iteration step $k$, the CBA parameter $\omega^k$ is fixed and we formulate a one-step stochastic gradient descent (SGD) to update the classifier network parameter $\theta^k$ in Eq.~(\ref{eq:inner-loss}), which can be represented as
\begin{equation}
    \theta^{k+1}(\omega) = 
    \theta^{k} - \alpha \cdot \nabla_{\theta} \mathcal{L}^{trn} \left(\mathcal{B}^{trn}; \mathcal F_{\theta^k, \omega^k} \right),
\label{eq:inner-opt}
\end{equation}
where $\alpha$ is the inner-loop learning rate.

(2) \textbf{Update $\omega$.}
After the one-step updating of $\theta^k$, we have obtained $\theta^{k+1}(\omega)$ which is a function of $\omega$.
Then we can optimize the $\omega$ in Eq.~(\ref{eq:outer-loss}) based on the updated $\theta^{k+1}(\omega)$ as following:
\begin{equation}
    \omega^{k+1} = 
    \omega^{k} - \beta \cdot \nabla_{\omega} \mathcal{L}^{buf} \left( \mathcal{B}^{buf}; f_{\theta^{k+1}(\omega)} \right),
\label{eq:outer-opt}
\end{equation}
where $\beta$ is the outer-loop learning rate.
Note that $\nabla_{\omega} \mathcal{L}^{buf}$ in Eq.~(\ref{eq:outer-opt}) introduces a second-order derivative, which can be easily implemented by the automatic differentiation system like Pytorch~\cite{pytorch}, and the detailed calculation can be found in Appendix A.
To alleviate the calculation burden of this derivation, we assume that $\omega$ is only correlated with the parameters of the linear classification layer. This allows us to only unroll the second-order derivation of the linear classification layer in Eq.~(\ref{eq:outer-opt}). As the linear classification layer involves a small number of parameters compared to the whole classifier network, our proposed algorithm is more efficient than other bi-level optimization algorithms \cite{MAML, DARTS, learn2reweight, meta-weight-net}.
A comprehensive discussion concerning computation and GPU memory utilization of our method is presented in Appendix D.4.
The complete training process is detailed in Alg.~\ref{alg:bac-training} and more intrinsic discussion can be found in the subsequent paragraph.

\begin{algorithm}[t]
\caption{Training of CBA in Online CL} 
\label{alg:bac-training}

    \textbf{Input:} new incoming sample batch $\mathcal{B}^t$, memory buffer $\mathcal{M}$ \\
    \textbf{Output:} classifier network parameter $\theta$, continual bias adaptor (CBA) parameter $\omega$
	\begin{algorithmic}[1]
        \State Initialize all network parameters as $\{\theta^0, \omega^0\}$.
        
        \While{$\mathcal{B}^t \neq \emptyset$}
        \Statex \qquad \textcolor{gray}{\# Inner-loop optimization:}
        \State $\mathcal{B}^{trn} = \mathcal{B}^t \cup \mathcal{B}^{buf}, \mathcal{B}^{buf} \subset \mathcal{M}$  \textcolor{gray}{\# Inner training data}
        \State Compute the inner-loop loss $\mathcal{L}^{trn}$ by Eq.~(\ref{eq:inner-loss})
        \State Update classifier network parameters $\theta$ by Eq.~(\ref{eq:inner-opt})
        \Statex \qquad \textcolor{gray}{\# Outer-loop optimization:}
        \State $\mathcal{B}^{buf} \subset \mathcal{M}$  \textcolor{gray}{\# Outer training data}
        \State Compute the outer-loop loss $\mathcal{L}^{buf}$ by Eq.~(\ref{eq:outer-loss})
        \State Update CBA parameters $\omega$ by Eq.~(\ref{eq:outer-opt}) 
        \EndWhile
	\end{algorithmic} 
\end{algorithm}

\subsection{Theoretical Analysis}
In this section, we theoretically analyze the proposed CBA model and the bi-level optimization procedure.
The following theorem claims that our algorithm inherently establishes gradient alignment between the loss on the corresponding training set $\mathcal{B}^{trn}$ and the memory buffer $\mathcal{B}^{buf}$.

\begin{thm}
Assume that the outer-loop loss $\mathcal{L}^{buf}(\cdot; f_\theta)$ is $\eta$ gradient Lipschitz continuous, then the bi-level optimization Eq.~(\ref{eq:inner-loss}) and Eq.~(\ref{eq:outer-loss}) potentially guarantees an alignment between the gradient of the outer-loop loss and the inner-loop loss with respect to the classification model parameter $\theta$, that is,
\label{thm:gradient_alignment}
\end{thm}
\vspace{-5mm}
\begin{sequation}
\begin{split}
    & \left\langle
        \frac{\partial \mathcal{L}^{buf} \left(\mathcal{B}^{buf}; f_{\theta^k(\omega)} \right)}{\partial \theta^k}
        \right., \left.
        \frac{\partial \mathcal{L}^{trn} \left(\mathcal{B}^{trn}; \mathcal{F}_{\theta^k, \omega} \right)}{\partial \theta^k}
    \right\rangle \\
    &\geq
    \frac{\alpha \eta}{2}
    \left\|
        \frac{\partial \mathcal{L}^{trn} \left(\mathcal{B}^{trn}; \mathcal{F}_{\theta^k, \omega} \right)}{\partial \theta^k}
    \right\|_2^2,
\end{split}
\label{eq:gradient_alignment}
\end{sequation}
\textit{where $\alpha > 0$ is the inner-loop learning rate and $\eta > 0$ is the Lipschitz constant.}

The detailed proof of Theorem~\ref{thm:gradient_alignment} is shown in Appendix B. Furthermore, this theorem reveals two insights into our algorithm.
On the one hand, this theorem explains why the CBA module adaptively assimilates the task-recency bias.
We hope the angle between the gradient w.r.t the classifier network parameter $\theta$ on the training set \textbf{with} CBA and the gradient on the memory buffer \textbf{without} CBA is as small as possible.
This means that the classifier network is expected to perform well on a balanced test set without the help of CBA, and the CBA only works during training to absorb the bias.
On the other hand, Theorem~\ref{thm:gradient_alignment} also shows why our method can mitigate forgetting effectively.
From Eq.~(\ref{eq:gradient_alignment}), our method is potentially close to some previous gradient-alignment-based CL works~\cite{AGEM, GEM}, which have already verified that the gradient alignment between the training set and the memory buffer aids the model in avoid forgetting.
Unfortunately, they do not take into account the negative impact of recency bias on the model.
Intuitively, our proposed method achieves more accurate gradient alignment because CBA can prevent bias from disturbing the gradient of the training set.

\section{Experiments}
\label{sec:experiments}
To validate the effectiveness of the proposed method, we compare our CBA to the state-of-the-art approaches on various datasets under online CL, blurry tasks, and offline CL settings. Moreover, for better comprehension of CBA, we also conduct extensive ablation experiments to analyze different components of our approach.
\begin{table*}[t]
\begin{small}
\begin{center}
\resizebox{1.0\textwidth}{!}{
\begin{tabular}{ccccccccccccc}
\toprule
\multirow{3}{*}{Method} & \multicolumn{4}{c}{Split CIFAR-10}                                       & \multicolumn{4}{c}{Split CIFAR-100}                                      & \multicolumn{4}{c}{Split Tiny-ImageNet}                                       \\ \cline{2-13} 
                        & \multicolumn{2}{c}{M = 0.2k}    & \multicolumn{2}{c}{M = 0.5k}    & \multicolumn{2}{c}{M = 2k}      & \multicolumn{2}{c}{M = 5k}      & \multicolumn{2}{c}{M = 2k}      & \multicolumn{2}{c}{M = 5k}      \\ \cline{2-13} 
                        & ACC $\uparrow$    & FM $\downarrow$  & ACC $\uparrow$    & FM $\downarrow$  & ACC $\uparrow$    & FM $\downarrow$  & ACC $\uparrow$    & FM $\downarrow$  & ACC $\uparrow$    & FM $\downarrow$  & ACC $\uparrow$    & FM $\downarrow$  \\
\midrule
iCaRL~\cite{icarl}      & 40.99          & 26.84          & 44.50          & 24.87          & 9.13           & 7.79           & 9.13           & 8.14           & 4.03           & \textbf{4.93}  & 4.03           & \textbf{5.15}  \\
LUCIR~\cite{LUCIR}      & 23.59          & 35.59          & 24.63          & 31.89          & 8.28           & 16.07          & 12.31          & 14.02          & 4.47           & 20.40          & 5.29           & 20.28          \\
BiC~\cite{bic}          & 27.71          & 66.45          & 35.47          & 47.92          & 16.32          & 36.70          & 20.89          & 32.33          & 5.43           & 40.14          & 7.50           & 38.52          \\ 
ER-ACE~\cite{online-er-ace} & 41.49          & 20.84          & 46.35          & 18.98          & 24.95          & \textbf{7.67} & 26.54          & \textbf{7.25} & 17.89           & 7.04          & 19.04           & 6.90          \\ 
SS-IL~\cite{SSIL}       & 37.92          & 15.64          & 41.22          & 11.46          & 24.90          & 9.85          & 25.60          & 10.23          & 17.91           & 7.93          & 18.53           & 8.26          \\ \hline
ER                      & 35.21          & 50.28          & 42.32          & 40.80          & 20.84          & 35.88          & 22.73          & 33.92          & 14.39          & 32.59          & 17.02          & 31.02          \\
\rowcolor{gray!40} ER-CBA (ours) & 37.27          & 41.39          & 45.41          & 29.36          & 25.67          & 10.21          & 27.59          & 8.74           & 18.06          & 13.16          & 20.20          & 10.16 \\
Gains                   & \textcolor{red}{+ 2.06} & \textcolor{red}{- 8.89} & \textcolor{red}{+ 3.09} & \textcolor{red}{-11.44} & \textcolor{red}{+ 4.83} & \textcolor{red}{-25.67} & \textcolor{red}{+ 4.86} & \textcolor{red}{-25.18} & \textcolor{red}{+ 3.67} & \textcolor{red}{-19.43} & \textcolor{red}{+ 3.18} & \textcolor{red}{-20.86} \\ \hline
DER++~\cite{DER}        & 40.17          & 41.84          & 43.44          & 40.63          & 16.87          & 44.46          & 17.61          & 44.53          & 11.81          & 39.88          & 12.31          & 39.93          \\
\rowcolor{gray!40} DER-CBA (ours) & 45.14          & 23.39          & 48.44          & \textbf{16.75} & 26.10          & 13.01          & 26.47          & 12.88          & 17.91          & 12.34          & 19.90          & 12.06 \\
Gains                   & \textcolor{red}{+ 4.97} & \textcolor{red}{-18.45} & \textcolor{red}{+ 5.00} & \textcolor{red}{-23.88} & \textcolor{red}{+ 9.23} & \textcolor{red}{-31.45} & \textcolor{red}{+ 8.86} & \textcolor{red}{-31.65} & \textcolor{red}{+ 6.10} & \textcolor{red}{-27.54} & \textcolor{red}{+ 7.59} & \textcolor{red}{-27.87} \\ \hline
RAR~\cite{RAR}          & 40.25          & 40.43          & 45.57          & 36.16          & 14.64          & 45.54          & 14.57          & 47.02          & 10.28          & 40.07          & 10.39          & 40.56          \\
\rowcolor{gray!40} RAR-CBA (ours)        & 43.28          & \textbf{20.10} & 48.45          & 17.42          & 23.87          & 13.10          & 24.19          & 14.00          & 16.70          & 11.48          & 17.29          & 12.56 \\ 
Gains                   & \textcolor{red}{+ 3.03} & \textcolor{red}{-20.33} & \textcolor{red}{+ 2.88} & \textcolor{red}{-18.74} & \textcolor{red}{+ 9.23} & \textcolor{red}{-32.44} & \textcolor{red}{+ 9.62} & \textcolor{red}{-33.02} & \textcolor{red}{+ 6.42} & \textcolor{red}{-38.59} & \textcolor{red}{+ 6.90} & \textcolor{red}{-28.00} \\ \hline
CLSER~\cite{CLSER}      & 42.01          & 42.33          & 44.48          & 36.83          & 22.48          & 34.80          & 23.18          & 34.35          & 15.34          & 32.48          & 17.28          & 31.79          \\
\rowcolor{gray!40} CLSER-CBA (ours)      & \textbf{44.31} & 27.55          & \textbf{49.63} & 19.99          & \textbf{26.90} & 9.41           & \textbf{29.09} & 8.05           & \textbf{19.31} & 12.80          & 
\textbf{21.62} & 9.67 \\
Gains                   & \textcolor{red}{+ 2.30} & \textcolor{red}{-14.78} & \textcolor{red}{+ 5.15} & \textcolor{red}{-16.84} & \textcolor{red}{+ 4.42} & \textcolor{red}{-25.39} & \textcolor{red}{+ 5.91} & \textcolor{red}{-26.30} & \textcolor{red}{+ 3.97} & \textcolor{red}{-19.68} & \textcolor{red}{+ 4.34} & \textcolor{red}{-22.12} \\
\bottomrule
\end{tabular}
}
\end{center}
\end{small}
\caption{Main results (ACC, higher is better, and FM, lower is better) on the three datasets with different memory buffer sizes. Our method applied on 4 baselines is shown with gray cells. \textbf{Bold} means the best results in all comparison methods and the gains of our method comparing the corresponding baselines are shown in \textcolor{red}{red} color.}
\label{tab:main-results}
\end{table*}

\begin{table*}[]
\begin{center}
\begin{small}
\scalebox{0.95}{
\begin{tabular}{ccccccccccccc}
\hline
\multirow{3}{*}{\thead{Split CIFAR10 \\ ($M=0.5k$)}}      & Method        & \multicolumn{2}{c}{$a_{1,5}$} & \multicolumn{2}{c}{$a_{2,5}$} & \multicolumn{2}{c}{$a_{3,5}$} & \multicolumn{2}{c}{$a_{4,5}$} & \multicolumn{2}{c}{$a_{5,5}$} & ACC    \\ \cline{2-13} 
        & ER                               & \multicolumn{2}{c}{28.27}   & \multicolumn{2}{c}{26.12}   & \multicolumn{2}{c}{31.95}   & \multicolumn{2}{c}{46.81}   & \multicolumn{2}{c}{78.48}   & 42.32  \\
        & \cellcolor{gray!40} ER-CBA (ours) & \multicolumn{2}{c}{\cellcolor{gray!40} 44.29}   & \multicolumn{2}{c}{\cellcolor{gray!40} 30.40}   & \multicolumn{2}{c}{\cellcolor{gray!40} 37.41}   & \multicolumn{2}{c}{\cellcolor{gray!40} 48.74}   & \multicolumn{2}{c}{\cellcolor{gray!40} 66.23}   & \cellcolor{gray!40} 45.41  \\ \hline
\multirow{3}{*}{\thead{Split CIFAR100 \\ ($M=5k$)}}     & Method        & $a_{1,10}$     & $a_{2,10}$     & $a_{3,10}$     & $a_{4,10}$     & $a_{5,10}$     & $a_{6,10}$     & $a_{7,10}$     & $a_{8,10}$     & $a_{9,10}$     & $a_{10,10}$    & ACC    \\ \cline{2-13} 
        & ER                               & 19.26        & 19.21        & 22.84        & 13.35        & 17.01        & 20.58        & 21.91        & 14.92        & 14.20        & 64.01        & 22.72 \\
        & \cellcolor{gray!40} ER-CBA (ours) & \cellcolor{gray!40} 25.92 & \cellcolor{gray!40} 22.35 & \cellcolor{gray!40} 34.82 & \cellcolor{gray!40} 27.81 & \cellcolor{gray!40} 28.44 & \cellcolor{gray!40} 35.77 & \cellcolor{gray!40} 32.83 & \cellcolor{gray!40} 30.04 & \cellcolor{gray!40} 17.83 & \cellcolor{gray!40} 20.05 & \cellcolor{gray!40} 27.58 \\ \hline
\multirow{3}{*}{\thead{Split Tiny-ImageNet \\ ($M=5k$)}} & Method        & $a_{1,10}$     & $a_{2,10}$     & $a_{3,10}$     & $a_{4,10}$     & $a_{5,10}$     & $a_{6,10}$     & $a_{7,10}$     & $a_{8,10}$     & $a_{9,10}$     & $a_{10,10}$    & ACC    \\ \cline{2-13} 
        & ER                               & 14.38        & 14.44        & 15.39        & 18.29        & 15.53        & 13.14        & 11.25        & 11.87        & 4.80         & 51.15        & 17.02 \\
        & \cellcolor{gray!40} ER-CBA (ours) & \cellcolor{gray!40} 21.58 & \cellcolor{gray!40} 18.84 & \cellcolor{gray!40} 24.44 & \cellcolor{gray!40} 25.30 & \cellcolor{gray!40} 20.79 & \cellcolor{gray!40} 18.26 & \cellcolor{gray!40} 18.83 & \cellcolor{gray!40} 18.89 & \cellcolor{gray!40} 11.33 & \cellcolor{gray!40} 23.75 & \cellcolor{gray!40} 20.20 \\ \hline
\end{tabular}}
\end{small}
\end{center}
\caption{Each task accuracy after the final task training $a_{t,T} (t = 1, \cdots, T)$ of ER and our ER-CBA on the three datasets. For Split CIFAR-10, task 5 is the new task, while task 10 is the incoming new task for Split CIFAR-100 and Split Tiny-ImageNet.}
\label{tab:per-task-acc}
\end{table*}

\begin{figure*}
\begin{center}
\subfloat[Prediction Task Distribution of ER and ER-CBA.]{
    \includegraphics[height=4cm]{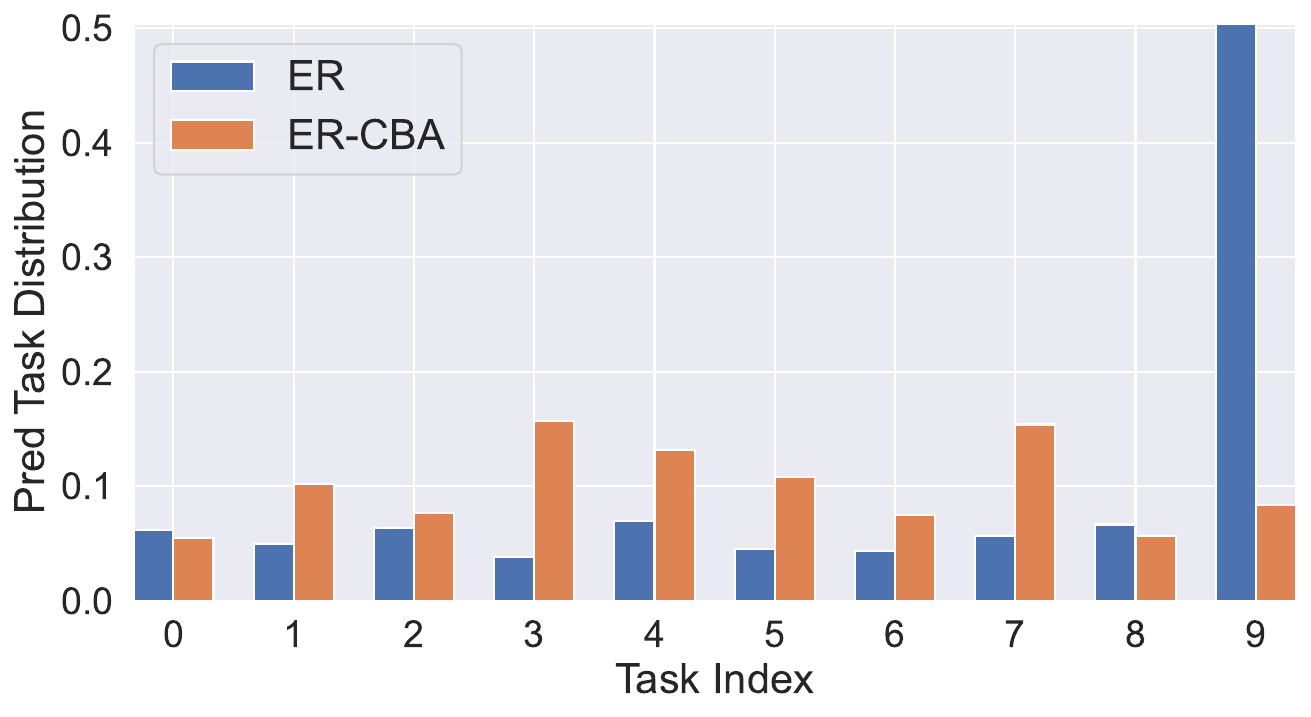}
}
\hspace{2pt}
\subfloat[Confusion Matrix of ER.]{
    \includegraphics[height=4cm]{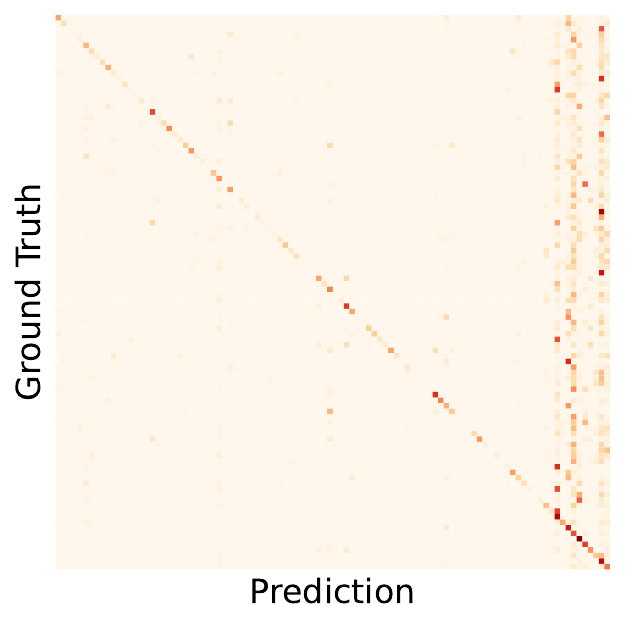}
}
\hspace{2pt}
\subfloat[Confusion Matrix of ER-CBA.]{
    \includegraphics[height=4cm]{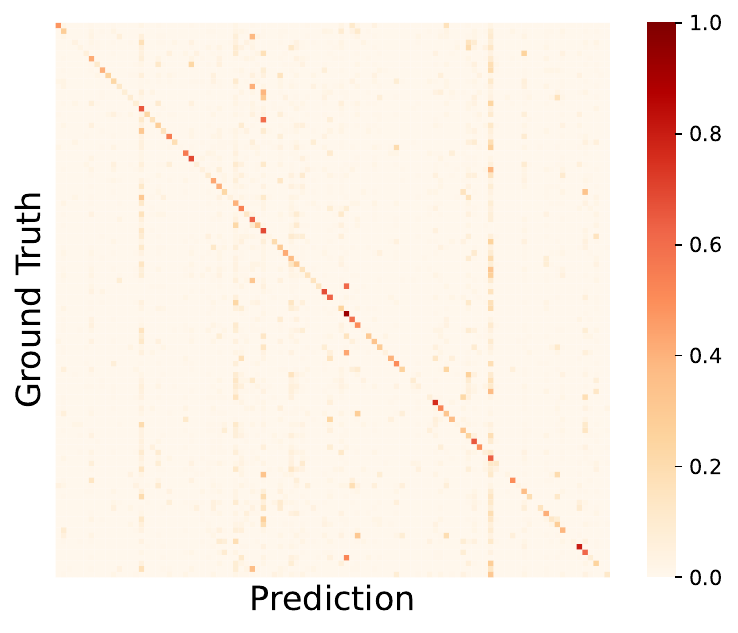}
 }
\end{center}
\vspace{-2mm}
\caption{(a): The prediction distribution of each task on Split CIFAR-100 with buffer size $M = 5k$. (b) and (c): The normalized confusion matrix of ER and ER-CBA, respectively.}
\vspace{-3mm}
\label{fig:task-distrib}
\end{figure*}

\subsection{Experimental Settings}
\label{subsec:exp-settings}
\medskip
\noindent\textbf{Experimental datasets.}
Following \cite{DER}, we choose three common datasets in CL, the \textbf{Split CIFAR-10} and the longer task sequence \textbf{Split CIFAR-100} and \textbf{Split Tiny-ImageNet}. Concretely, Split CIFAR-10 contains five binary classification tasks, which are constructed by evenly splitting ten classes of CIFAR-10~\cite{cifar}. Split CIFAR-100 and Split Tiny-ImageNet both include ten disjoint tasks, each of which has 10 and 20 classes, respectively (see Appendix C for details of the three datasets).

\medskip
\noindent\textbf{Evaluation metrics.} To comprehensively evaluate all comparison methods, we consider the following metrics:
\begin{itemize}
    \item \textbf{Average Accuracy} (ACC $\uparrow$): calculate the average accuracy of the model trained on all tasks, \textit{i.e.} ${\rm{ACC}} = 1/T \sum^{T}_{t=1} a_{t, T}$, where $a_{i, j}$ represents the accuracy of the task $i$ after training on the task $j$.
    
    \item \textbf{Forgetting Measure} (FM $\downarrow$): average of the decreasing from the best accuracy to the final accuracy, \textit{i.e.} ${\rm{FM}} = 1/T \sum^{T}_{t=1} a^*_t - a_{t, T}$, where the $a^*_t$ is the best accuracy of task $t$ in the whole training process.
    
    \item \textbf{Area Under the Curve of Accuracy} (${\rm{{ACC}_{AUC}}} \uparrow$): this metric is the area under the curve of the accuracy~\cite{ACC_AUC}, \textit{i.e.,} ${\rm{{ACC}_{AUC}}} = \sum_i \bar{a}(i\cdot \triangle n)\cdot \triangle n$, where $\bar{a}(i)$ represent the average accuracy when the model training at step $i$, and $\triangle n$ is the interval training step which is 5 for faster evaluation in our experiments.
\end{itemize}

\begin{figure}
\begin{center}
   \includegraphics[width=0.7\columnwidth]{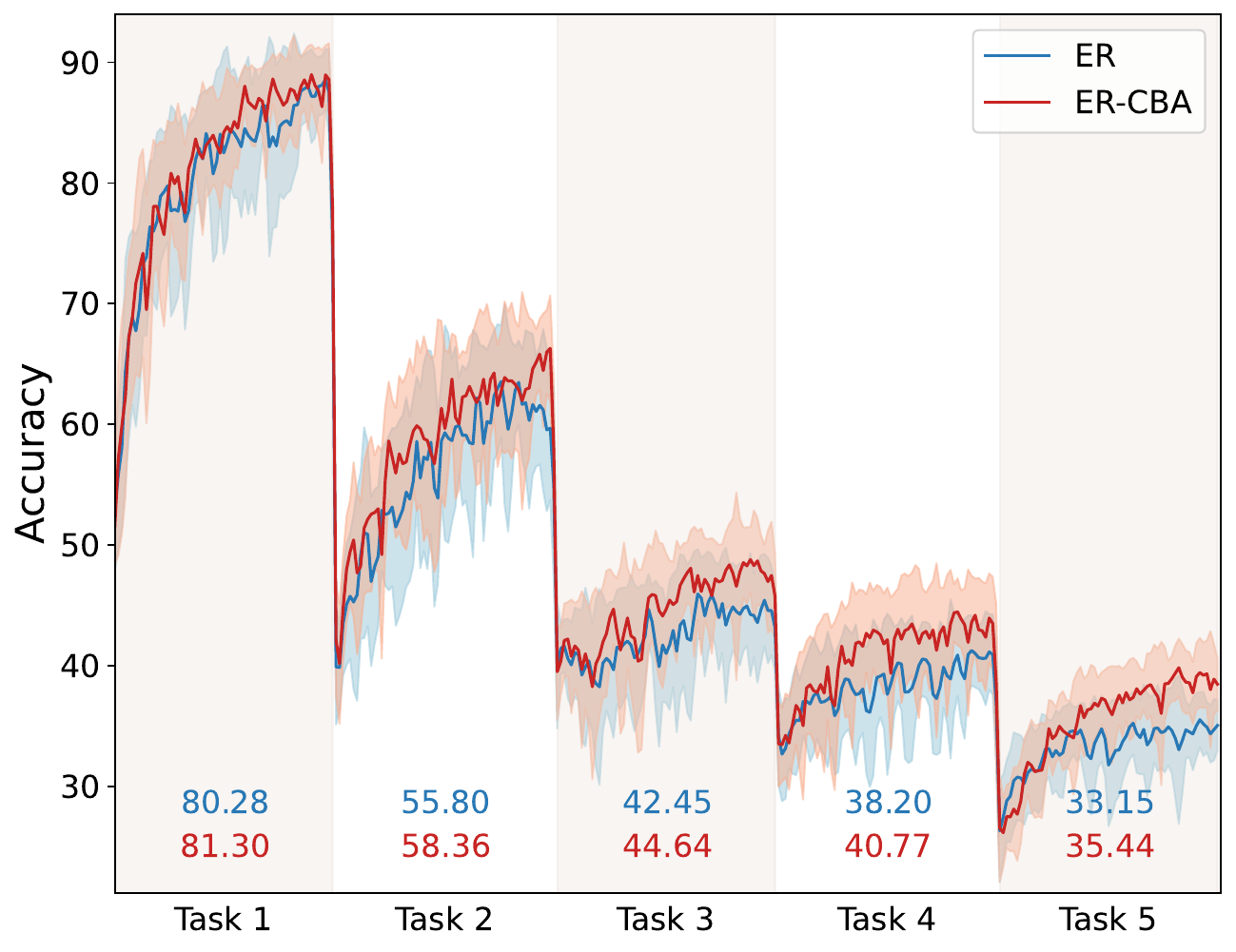}
\end{center}
\vspace{-2mm}
\caption{The average accuracy in the whole training process. The $\rm{{ACC}_{AUC}}$ in each stage is shown in the figure, where the 1st row is the $\rm{{ACC}_{AUC}}$ of the baseline ER and the 2nd row is that of our method ER-CBA.}
\vspace{-3mm}
\label{fig:acc_auc}
\end{figure}

\medskip
\noindent\textbf{Implementation details.} We adopt the commonly used ResNet-18 as our backbone~\cite{CLSER, DER, bic}, and train all methods using the Stochastic Gradient Descent (SGD) optimizer. For our CBA, we set the learning rate as 0.001 for Split CIFAR-10, and 0.01 for Split CIFAR-100 and Split Tiny-ImageNet. To verify our method consistently outperforms other baselines, each reported result in the online CL setting is an average of 10 repeated runs, and each result in the offline is an average of 5 runs. More details about the baselines and implementation are listed in Appendix C.

\subsection{Comparison on Disjoint Scenario}
\label{subsec:main-results}
\medskip
\noindent\textbf{CBA enhanced current rehearsal-based methods.}
We first investigate the performance of CBA by plugging it into ER, DER, RAR, and CLSER. These four baselines adopt different replay strategies without the correction of the distribution shift. Table~\ref{tab:main-results} shows that our proposed CBA improves the ACC of all these four rehearsal-based methods by up to 9.23\% on these three widely used disjoint CL benchmarks. It can also be observed that the Forgetting Measure is reduced by up to 33.02\%, indicating the CBA can significantly alleviate catastrophic forgetting. Additionally, our method achieves consistent improvement for the comparison methods under different memory buffer sizes, which verifies that our method has strong generalizability.
Further results can be found in Appendix D.

To investigate the changing of average accuracy throughout the training process, we display the accuracy of the baseline ER and our ER-CBA in Fig.~\ref{fig:acc_auc}, which is evaluated on Split CIFAR-10 ($M=0.2k$).
As we can see, our method can surpass the baseline within just a few training iteration steps. In this figure, $\rm{{ACC}_{AUC}}$ of each task is also calculated, revealing a consistent improvement by our method over the baseline ER.
Moreover, our method maintains robust performance improvement even when constrained by small memory buffer sizes for each dataset, and further results are shown in Appendix D.

\medskip
\noindent\textbf{Analysis of other baselines}
Note that LUCIR employs a weight normalization prior, and BiC designs an additional linear layer, both attempting to rectify the distribution shift from a class imbalance view. However, they oversimplified the distribution shift issue, leading to suboptimal performance across these three benchmarks.
iCaRL fails on the larger datasets Split CIFAR-100/Tiny-ImageNet, suggesting that the NCM classifier relies on accurate and well-separate class means, which is difficult to obtain in online CL on large datasets.
Because of the worse accuracy of iCaRL on these two larger datasets, the lower FM in Table~\ref{tab:main-results} does not make any sense.
As for ER-ACE and SS-IL, although they exhibit decent ACC performance, they over-suppress the accuracy of new classes during training, which caused the lower FM metric.

\subsection{Comparison on Blurry Scenario}
\label{subsec:task-blurry}
Following~\cite{gradient-selection-blurry, rainbow-blurry}, we adopt the \textit{Blurry-$K$} online CL setting. It simulates a practical situation where task boundaries are unclear, causing class overlaps across all tasks. Specifically, a fraction ($K\%$) of the training data from one task may appear in other tasks. Here we set $K=10$ as an illustration, and the comparative results are summarized in Table~\ref{tab:blurry-results}.

It can be observed from Table~\ref{tab:blurry-results} that our CBA module can vastly improve the performance of the corresponding baselines. For example, the ACC of CLSER-CBA is higher than the baseline CLSER by about 3.19\% and 5.17\% with memory buffer sizes $M\!=\!0.2k$ and $0.5k$, respectively. And the FM of CLSER-CBA is lower than it of CLSER by about 8.46\% and 14.97\% with $M\!=\!0.2k$ and $0.5k$, respectively. It shows that the proposed CBA module can be flexibly adapted to the classifier network without being perturbed by ambiguous task boundaries seriously, which also verifies the strength of our method.

\subsection{Comparison under the Offline CL }
\label{subsec:offline-CL}
To further verify the effectiveness and strength of our method, we extend our method to the offline CL setting, where each task can be trained over multiple epochs. As depicted in Table~\ref{tab:offline-results}, these baselines perform better in the offline CL compared to the online context. Nevertheless, our method can also improve the corresponding baselines by correcting the distribution shift online.
It is worth noting that CBA improves the ER significantly and makes it comparable with the other three baselines, which further illustrates the negative impact of task-recency bias on CL model performance and the strength of our method.

\begin{table}[]
\begin{small}
\begin{center}
\begin{tabular}{ccccc}
\hline
\multirow{3}{*}{Method} & \multicolumn{4}{c}{Split CIFAR-10}                             \\ \cline{2-5} 
                        & \multicolumn{2}{c}{$M = 0.2k$} & \multicolumn{2}{c}{$M = 0.5k$} \\ \cline{2-5} 
                        & ACC $\uparrow$ & FM $\downarrow$ & ACC $\uparrow$ & FM $\downarrow$ \\ \hline
ER                                  & 44.15          & 40.33          & 50.37          & 30.29         \\
\rowcolor{gray!40} ER-CBA (ours)    & 48.42          & 30.60          & 52.26          & 20.72         \\ 
Gains                               & \textcolor{red}{+ 4.27} & \textcolor{red}{- 9.73} & \textcolor{red}{+ 1.89} & \textcolor{red}{- 9.57} \\ \hline
DER++~\cite{DER}                    & 49.25          & 31.50          & 51.81          & 33.28          \\
\rowcolor{gray!40} DER-CBA (ours)   & 51.38          & 23.75          & 52.11          & 17.13          \\  
Gains                               & \textcolor{red}{+ 2.13} & \textcolor{red}{- 7.75} & \textcolor{red}{+ 0.30} & \textcolor{red}{-16.15} \\ \hline
RAR~\cite{RAR}                      & 47.20          & 37.62          & 48.46          & 37.78          \\
\rowcolor{gray!40} RAR-CBA (ours)   & \textbf{51.69} & \textbf{18.66} & 52.25          & \textbf{12.35} \\  
Gains                               & \textcolor{red}{+ 4.49} & \textcolor{red}{-18.96} & \textcolor{red}{+ 3.79} & \textcolor{red}{-25.43} \\ \hline
CLSER~\cite{CLSER}                  & 47.57          & 35.77          & 49.52          & 33.54          \\
\rowcolor{gray!40} CLSER-CBA (ours) & 50.76          & 27.31          & \textbf{54.69} & 18.57          \\  
Gains                               & \textcolor{red}{+ 3.19} & \textcolor{red}{- 8.45} & \textcolor{red}{+ 5.17} & \textcolor{red}{-14.97} \\ \hline
\end{tabular}
\end{center}
\end{small}
\vspace{-2mm}
\caption{ACC and FM of our method applied on 4 baselines under the `Blurry-10' settings on Split CIFAR-10.}
\label{tab:blurry-results}
\end{table}

\begin{table}[]
\begin{small}
\begin{center}
\begin{tabular}{ccccc}
\hline
\multirow{3}{*}{Method} & \multicolumn{4}{c}{Split CIFAR-10}                                       \\ \cline{2-5} 
                        & \multicolumn{2}{c}{$M=0.2k$}      & \multicolumn{2}{c}{$M=0.5k$}      \\ \cline{2-5} 
                        & ACC $\uparrow$ & FM $\downarrow$ & ACC $\uparrow$ & FM $\downarrow$ \\ \hline
ER                                  & 55.80          & 47.77          & 66.06          & 33.02          \\
\rowcolor{gray!40} ER-CBA (ours)    & 64.12          & 31.40          & 72.86          & 20.63          \\ 
Gains                               & \textcolor{red}{+ 8.32} & \textcolor{red}{-16.37} & \textcolor{red}{+ 6.80} & \textcolor{red}{-12.39} \\ \hline
DER++~\cite{DER}                    & 63.33          & 35.83          & 72.78          & 23.40          \\
\rowcolor{gray!40} DER-CBA (ours)   & 65.64          & 31.78          & 74.37          & 20.06          \\ 
Gains                               & \textcolor{red}{+ 2.31} & \textcolor{red}{- 4.05} & \textcolor{red}{+ 1.59} & \textcolor{red}{- 3.34} \\ \hline
RAR~\cite{RAR}                      & 60.57          & 39.44          & 69.73          & 26.99          \\
\rowcolor{gray!40} RAR-CBA (ours)   & 63.49          & 32.79          & 71.55          & 23.21          \\ 
Gains                               & \textcolor{red}{+ 2.92} & \textcolor{red}{- 6.65} & \textcolor{red}{+ 1.82} & \textcolor{red}{- 3.78} \\ \hline
CLSER~\cite{CLSER}                  & 65.73          & 30.32          & 73.45          & 19.45          \\
\rowcolor{gray!40} CLSER-CBA (ours) & \textbf{67.40} & \textbf{27.16} & \textbf{74.51} & \textbf{18.75} \\ 
Gains                               & \textcolor{red}{+ 1.67} & \textcolor{red}{- 3.16} & \textcolor{red}{+ 1.06} & \textcolor{red}{- 0.70} \\ \hline
\end{tabular}
\end{center}
\end{small}
\vspace{-2mm}
\caption{ACC and FM of our method applied on 4 baselines under the offline CL settings on Split CIFAR-10.}
\vspace{+2mm}
\label{tab:offline-results}
\end{table}

\begin{table}[]
\begin{small}
\begin{center}
\scalebox{0.9}{
\begin{tabular}{cccccc}
\hline
\multirow{3}{*}{Method} & \multirow{3}{*}{Details} & \multicolumn{4}{c}{Split CIFAR-10} \\ \cline{3-6} 
                        &                          & \multicolumn{2}{c}{$M=0.2k$}      & \multicolumn{2}{c}{$M=0.5k$}      \\ \cline{3-6} 
                        &                          & ACC $\uparrow$ & FM $\downarrow$ & ACC $\uparrow$ & FM $\downarrow$ \\ \hline
ER     & -                   & 35.21          & 50.28          & 42.32          & 40.80          \\
ER-CBA & $\mathcal{L}^{trn}$ & 31.95          & 51.76          & 36.66          & 44.55          \\
ER-CBA & $\mathcal{L}^{trn} + \mathcal{L}^{buf}$ & 33.35          & 52.98          & 44.58          & 35.95          \\
ER-CBA & 64 hidden units     & 33.50          & 44.44          & 37.48          & 37.82          \\
ER-CBA & 1024 hidden units   & \textbf{38.46} & \textbf{42.73} & \textbf{46.85} & \textbf{24.86} \\
ER-CBA & 4 layers MLP        & 37.57          & 41.81          & 45.23          & 27.91          \\
\rowcolor{gray!40} ER-CBA & ours                & 37.27          & 41.39          & 45.41          & 29.36          \\ \hline
\end{tabular}
}
\end{center}
\end{small}
\vspace{-2mm}
\caption{The results of ablation study.}
\vspace{-2mm}
\label{tab:ablation-results}
\end{table}

\subsection{Discussion and Ablation Study.}
\label{subsec:discussion}
To better understand the CBA, we conduct experiments and analyze the results to answer the following questions.

\medskip
\noindent\textbf{Question 1: Does CBA help the model adapt to distribution shift?}
To ascertain this, it is essential to investigate how the accuracy of each task changes after training on the final task, \textit{i.e.}, $a_{t, T} (t=1, \cdots, T)$.
In Table~\ref{tab:per-task-acc}, we exhibit $a_{t, T}$ and ACC of the baseline ER and ER-CBA on Split CIFAR-10 ($M=0.5k$), Split CIFAR-100 ($M=5k$), and Split Tiny-ImageNet ($M=5k$), respectively. It shows that the CBA module can significantly improve the performance of previous tasks and reduce forgetting of the model. Additionally, our CBA can effectively trade off the accuracy of old and new tasks, which indicates that CBA can prevent the model from being disturbed by distribution shifts (caused by incoming new tasks).

\medskip
\noindent\textbf{Question 2: How does the predictive distribution with and without the CBA module compare?}
To investigate this question, we calculate the prediction distribution of each task after training on the final task of Split CIFAR-100 ($M=5k$) in Fig.~\ref{fig:task-distrib} (a), and show the corresponding confusion matrix of ER and ER-CBA in Fig.~\ref{fig:task-distrib} (b) and (c), respectively.
Obviously, the baseline ER tends to predict test samples as new classes with high probability, which leads to significantly higher accuracy for new tasks than old tasks within ER.
In contrast, our method can suppress the prediction probability of new tasks and automatically adapt to the changing distribution.
Therefore, our method treats test samples from all seen classes more equally and achieves better overall performance, which reveals how CBA helps the model deal with the data distribution shift problem.

\medskip
\noindent\textbf{Question 3: How effective are the various components of the proposed model?}
We conduct a comprehensive ablation study to investigate the contribution of each key component of the proposed algorithm and the results on Split CIFAR-10 are shown in Table~\ref{tab:ablation-results}. Firstly, we verify the effectiveness of the bi-level optimization part. We replace the bi-level optimization procedure with an end-to-end training parallel, that is, optimize the parameters $\theta$ of the classification model and the CBA parameter $\omega$ together. The bi-level optimization and the end-to-end training correspond to 1) have the rehearsal training loss Eq.~(\ref{eq:inner-loss}) only; 2) sum the inner-loop training loss Eq.~(\ref{eq:inner-loss}) and the outer-loop training loss Eq.~(\ref{eq:outer-loss}).
It can be noted that the end-to-end training cannot help the model alleviate the distribution shift and even hurt the performance of baseline ER (the 2nd and 3rd rows in Table~\ref{tab:ablation-results}), which demonstrate that the bi-level optimization is essential for our model training.

We also explore the effect of different MLP architectures in the CBA module where we use a two-layer MLP with 256 hidden units.
Specifically, we test the performance of 1) a two-layer MLP with 64 and 1024 hidden units; 2) a four-layer MLP with 256 hidden units.
As shown in the 4-6th rows of Table~\ref{tab:ablation-results}, the representation ability of the MLP with only 64 hidden units is not enough to fit the bias, which leads to lower performance than the baseline. And the MLP with 1024 hidden units performs even better than 256 hidden units, suggesting that the stronger CBA architecture might perform better even though it will introduce more parameters and slow down the training process. 
Since the performance of the complicated four-layer MLP with 256 hidden units is almost the same as the two-layer MLP, we choose the two-layer MLP with 256 hidden units as the base architecture of our CBA.

\section{Conclusion}
\label{sec:conclusion}
In this paper, we tackle online CL from the perspective of modeling the posterior distribution shift arising from time-varying data streams. We propose a Continual Bias Adaptor (CBA) module, which aims to augment the original classifier network to adapt to catastrophic distribution change during training, such that the classifier network is capable of learning a stable consolidation of the previously learned knowledge across all tasks. In the inference stage, CBA can be removed and requires no additional calculation burden or memory overhead. The effectiveness of the proposed method is demonstrated both theoretically and empirically. In theory, we explain the reason why our method can significantly alleviate catastrophic forgetting in online CL. Empirical results show that the proposed algorithm can be applied to many rehearsal-based baselines and consistently improve their performance, which verifies that our method can effectively consolidate the learned knowledge.

\section*{Acknowledgement}
This research was supported by the National Key R\&D Program of China (2020YFA0713900), the China NSFC projects under contracts 61906144, 61721002, 12226004, and the Macao Science and Technology Development Fund under Grant 061/2020/A2.

{\small
\bibliographystyle{ieee_fullname}
\bibliography{egbib}
}

\end{document}